\documentclass[letterpaper]{article}
\usepackage{aaai}
\usepackage{times}
\usepackage{helvet}
\usepackage{courier}

\usepackage{enumitem}
\usepackage{graphicx}
\usepackage{psfrag}
\usepackage{calc}
\usepackage{url}
\usepackage{graphics}
\usepackage{amsmath}
\usepackage{amssymb}
\usepackage{relsize}
\usepackage{multirow}
\usepackage{booktabs}
\usepackage{tikz}
\usepackage{verbatim}
\usepackage{array}

\usepackage{graphicx}
\usepackage{multicol}
\usepackage[T1]{fontenc}

\begin{document}
%
\title{Inferring Robot Task Plans from Human Team Meetings: \\
A Generative Modeling Approach with Logic-Based Prior}
\author{Been Kim, Caleb M. Chacha \and Julie Shah\\
Massachusetts Institute of Technology\\
Cambridge, Masachusetts 02139\\
\{beenkim, c\_chacha, julie\_a\_shah\}@csail.mit.edu}

\maketitle
\begin{abstract}
\begin{quote}   
We aim to reduce the burden of programming and deploying autonomous systems to work in concert with people in time-critical domains, 
such as military field operations and disaster response. Deployment plans for these operations are frequently negotiated on-the-fly by teams of human planners. 
A human operator then translates the agreed upon plan into machine instructions for the robots. We present an algorithm that reduces this translation 
burden by inferring the final plan from a processed form of the human team's planning conversation. Our approach combines probabilistic generative modeling with 
logical plan validation used to compute a highly structured prior over possible plans. This hybrid approach enables us to overcome the challenge of 
performing inference over the large solution space with only a small amount of noisy data from the team planning session. We validate the algorithm 
through human subject experimentation and show we are able to infer a human team's final plan with 83\% accuracy on average. We also describe a 
robot demonstration in which two people plan and execute a first-response collaborative task with a PR2 robot. To the best of our knowledge, 
this is the first work that integrates a logical planning technique within a generative model to perform plan inference. 
\end{quote}
\end{abstract}

\section{Introduction}

Robots are increasingly introduced to work in concert with people in
high-intensity domains, such as military field operations and disaster response.
They have been deployed to gain access to areas that are inaccessible to people~\cite{casper2003human,micire2002analysis},
and to inform situation assessment~\cite{larochelle2011establishing}. The human-robot interface 
has long been identified as a major bottleneck in utilizing these robotic systems to their full potential~\cite{murphy2004human}.
As a result, we have seen significant research efforts aimed at easing the use of these systems in the field, including careful design and validation of supervisory and control interfaces~\cite{jones2002autonomous,Cummings:2007,BarnesCJR11,GoodrichEtal09J}.
Much of this prior work has focused on ease of use at
``execution time.'' However, a significant bottleneck also exists in planning the deployments of autonomous systems and in programming these systems to coordinate their task execution with
the rest of the human team. 
Deployment plans are frequently negotiated by human team members on-the-fly
and under time pressure ~\cite{casper2002workflow,casper2003human}. For a robot to execute part of this plan, a human operator must transcribe and translate the result of the team planning session.

In this paper we present an algorithm that reduces this translation 
burden by inferring the final plan from a processed form of the human team's planning conversation. Our approach combines probabilistic generative modeling with 
logical plan validation which is used to compute a highly structured prior over possible plans. This hybrid approach enables us to overcome the challenge of 
performing inference over the large solution space with only a small amount of noisy data from the team planning session. 

Rather than working with raw natural language, our 
algorithm takes as input a structured form of the human dialog. 
Processing human dialogue 
into more machine understandable forms is an important research area~\cite{kruijff2010continual,tellex11,koomen2005generalized,palmer2010semantic,pradhan2004shallow},
but we view this as a separate problem and do not focus on it in this paper.

The form of input we use preserves many of the 
challenging aspects of natural human planning conversations and can be thought of as very `noisy'
observations of the final plan.  Because the team is discussing the plan under time pressure, 
the planning sessions often consist of a small number of succinct communications.
Our approach reliably infers the majority of final agreed upon plan, despite the small amount of noisy data.

We validate the algorithm 
through human subject experimentation and show we are able to infer a human team's final plan with 83\% accuracy on average. We also describe a 
robot demonstration in which two people plan and execute a first-response collaborative task with a PR2 robot. To the best of our knowledge, 
this is the first work that integrates a logical planning technique within a generative model to perform plan inference.


\section{Problem Formulation}

Disaster response teams are increasingly utilizing web-based planning tools to plan their deployments~\cite{LincolnNICS}. 
Dozens to hundreds of responders log in to plan their deployment using audio/video conferencing, text chat, and annotatable maps. 
In this work, we focus on inferring the final plan using the text data that can be logged from chat or transcribed speech. This section 
formally describes our problem formulation, including the inputs and outputs of our inference algorithm.

\subsection{Planning Problem}

A plan consists of a set of actions together with execution timestamps associated with those actions. A plan is \emph{valid} if it achieves a user-specified goal state without
violating user-specified plan constraints. In our work, the full set of possible actions is not necessarily specified apriori and actions may also be derived from the dialog. 

Actions may be constrained to execute in $sequence$ or in $parallel$ with other actions.  
Other plan constraints include discrete resource constraints (e.g.~there are two medical teams), and temporal deadlines on time durative actions 
(e.g. a robot can be deployed for up to one hour at a time due to battery life constraints).  Formally, we assume our planning problem may be represented in Planning Domain Description Language (PDDL) 2.1.
 
\begin{figure}
 \begin{center}
  \includegraphics[scale=.3]{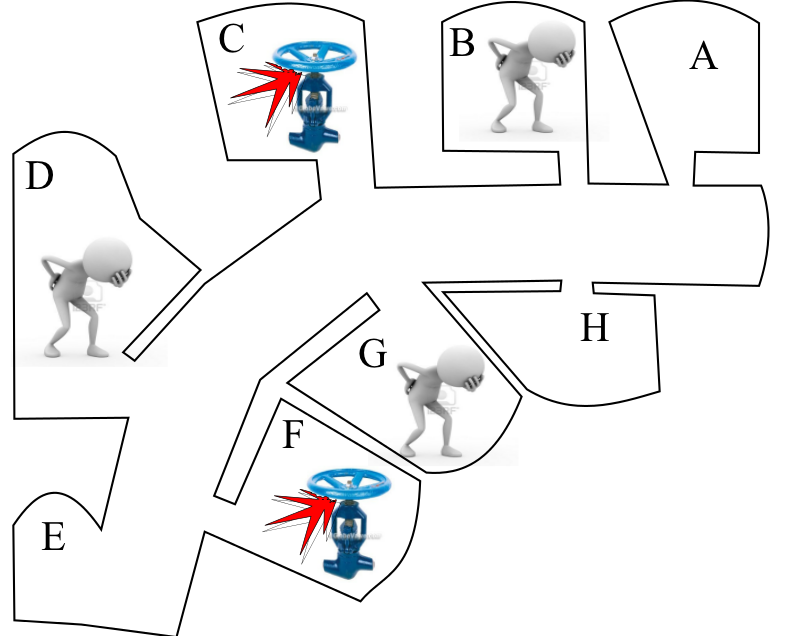}
\caption{Fictional rescue scenario\label{fig:scenario}}
 \end{center}
\end{figure}

In this paper, we consider a fictional rescue scenario that involves a radioactive material leakage accident
in a building with multiple rooms, as shown in Figure~\ref{fig:scenario}.
This is the scenario we use in human subject experiments for proof-of-concept of our approach.
A room either
has a patient that needs to be assessed in person or a valve that needs to be fixed. The scenario is specified by the following
information.

\noindent\textbf{Goal State:} All patients are assessed in person by a medical crew. All valves are fixed by a mechanic. All rooms are inspected by a robot.
 
\noindent\textbf{Constraints:} For safety, the radioactivity of a room must be inspected by a robot before human crews can be sent to the room (sequence constraint). There are two medical crews, red and blue 
(discrete resource constraint). There is one human mechanic (discrete resource constraint). There are two robots, red and blue (discrete resource constraint).

\noindent\textbf{Assumption:} All tasks (e.g.~inspecting a room, fixing a valve) take the same unit of time and there are no hard temporal constraints in the plan. 
This assumption was made to conduct the initial proof-of-concept experimentation described in this paper. 
However, the technical approach readily generalizes to variable task durations and temporal constraints, as described later.

The scenario produces a very large number of possible plans (more than $10^{12}$), many of which are 
valid for achieving the goals without violating the constraints.

We assume that the team reaches an agreement on a final plan. Techniques introduced by~\cite{Kim13Cogsima} can be used to detect the strength of
agreement, and encourage the team to discuss further to reach an agreement if necessary. We leave for future study the situation where the team agrees on a flexible plan with multiple options to be decided among later. 
While we assume that the team is more likely to agree on a valid plan, 
we do not rule out the possibility that the final plan is invalid. 



\subsection{Algorithm Input}
Text data from the human team conversation is collected in the form of utterances, where each utterance is  
one person's turn in the discussion, as shown in Table~\ref{tab:example}. The input to our algorithm
is a machine understandable form of human conversation data, as illustrated in the right-hand column of Table~\ref{tab:example}.
This structured form captures the actions discussed and the proposed ordering relations among actions for each utterance. 

Although we are not working with raw natural language, this form of data still captures many of the characteristics that make plan inference based on human conversation challenging.   
Table~\ref{tab:example} shows part of the data, using the following short-hand:
\begin{quote}
ST = SendTo \\
rr = red robot, br = blue robot \\
rm = red medical, bm = blue medical\\
e.g. ST(br, A) = SendTo(blue robot, room A)\\
\end{quote}
\begin{table}
\begin{center}
\small
\begin{tabular}{|p{.25cm}|p{3.8cm}|p{3.2cm}|}
\hline
& Natural dialogue & Structured form\\
\hline
\
U1 &  So I suggest using Red robot to cover ``upper'' rooms (A, B, C, D) and Blue robot to cover ``lower'' rooms (E, F, G, H). 
& (\{ST(rr,A),ST(br,E), ST(rr,B),ST(br,F), ST(rr,C),ST(br,G), ST(rr,D), ST(br,H)\}) \\

U2 & Okay. so first send Red robot to B and Blue robot to G?
& (\{ST(rr,B),ST(br,G)\})\\

U3 & Our order of inspection would be (B, C, D, A) for Red and then (G, F, E, H) for Blue.
& (\{ST(rr,B),ST(br,G)\}, \{ST(rr,C),ST(br,F)\}, \{ST(rr,D),ST(br,E)\}, \{ST(rr,A), ST(br,H)\}) \\
U4&  Oops I meant (B, D, C, A) for Red.
& (\{ST(rr,B)\},\{ST(rr,D)\}, \{ST(rr,C)\},\{ST(rr,A)\}) \\
& $\cdots$ & \\
U5& So we can have medical crew go to B when robot is inspecting C
&  (\{ST(m,B), ST(r,C)\})\\
& $\cdots$ & \\
U6&  First, Red robot inspects B
&  (\{ST(r,B)\})\\
U7& Yes, and then Red robot inspects D, Red medical crew to treat B
&  (\{ST(r,D),ST(rm,B)\})\\\hline
\end{tabular}
\caption{Dialogue and structured form examples\label{tab:example}}
\end{center}
\end{table}

\subsubsection{Utterance tagging}
An utterance is tagged as an \emph{ordered tuple of sets of grounded predicates}. 
Each predicate represents an action applied to a set of objects (crew member, robot, room, etc.),
as is the standard definition of a dynamic predicate in PDDL. Each set of grounded 
predicates represents a collection of actions that, according to the utterance, should happen in parallel.
The order of the sets of grounded predicates indicates the \emph{relative} order in which 
these collections of actions should happen. For example, (\{ST(rr, B), ST(br, G)\}, \{ST(rm, B)\})
corresponds to simultaneously sending the red robot to room B and the blue robot to room G, followed by 
sending the red medical team to room B. 

As one can see, the structured dialog is still very noisy.
Each utterance (i.e.~U1-U7) discusses a partial plan and only predicates that are explicitly mentioned
in the utterance are tagged (e.g.~U6-U7: the ``and then'' in U7 implies a sequencing constraint with the predicate discussed in U6, but the structured form of U7 does not include ST(r,B)).
Typos and misinformation are tagged as they are (e.g.~U3), and the utterances to revise information are not placed in context (e.g.~U4).
Utterances that clearly violate ordering constraints (e.g.~U1: all actions cannot happen at the same time) are tagged as they are. 
In addition, the information regarding whether the utterance was a suggestion, rejection or agreement of an partial plan is not
coded. 

Note that the utterance tagging only contains information about \emph{relative ordering} between the predicates appearing in 
that utterance, not the \emph{absolute ordering} of their appearance in the final plan.
For example, U2 specifies that the two grounded predicates happen at the same time.
It does not say when the two predicates happen in the final plan or whether other predicates happen in parallel.
This simulates how humans perceive the conversation --- at each utterance, humans only observe 
the relative ordering, and infer the absolute order of predicates based on the whole conversation 
and an understanding of which orderings make a valid plan.

\subsection{Algorithm Output}
The output of the algorithm is an inferred final plan, which is sampled from the probability distribution over final plans.
The final plan has a similar representation to the structured utterance tags, and is represented as an \emph{ordered tuple of sets of grounded predicates}.
The predicates in each set represent actions 
that happen in parallel, and the ordering of sets indicates sequence. Unlike the utterance tags, the sequence ordering relations in the final plan 
represent the \emph{absolute} order in which the actions are to be carried out. 
For example, a plan can be represented as
$(\{A_1, A_2\},\{A_3\}, \{A_4,A_5,A_6\})$, where $A_i$ represents a predicate.
In this plan, $A_1$ and $A_2$ will happen at plan time step $1$, $A_3$ happens at plan time step $2$, and so on.

 \section{Technical Approach and Related Work}

It is natural to take a probabilistic approach to the plan inference problem since we are working with noisy data. 
However the combination of a small amount of noisy data and a very large number of possible plans
means that an approach using typical uninformative priors over plans will frequently fail to converge to the team's plan in a timely manner.

This problem could also be approached as a logical constraint problem of partial order planning,
if there were no noise in the utterances. In other words, if the team discussed only the partial plans relating to the final plan and did not make any errors or revisions, 
then a plan generator such as a PDDL solver \cite{Coles20091} could produce the final plan with global sequencing.
Unfortunately human conversation data is sufficiently noisy to preclude this approach.

This motivates a combined approach. We build a probabilistic generative model for the structured utterance observations. We use a logic-based plan validator \cite{howey2004val} to compute 
a highly structured prior distribution over possible plans, which encodes our assumption that the final plan is likely, but not required, to be a valid plan. 
This combined approach naturally deals with 
noise in the data and the challenge of performing inference over plans with only a small amount of data. 
We perform sampling inference in the model using Gibbs sampling and Metropolis-Hastings steps within
to approximate the posterior distribution over final plans. 
We show through empirical validation with human subject experiments that the algorithm achieves 83\% accuracy on average.

Combining a logical approach with probabilistic modeling has gained interest
in recent years. 
\cite{getoor2011learning} introduce a language for describing statistical models over typed relational domains
and demonstrate model learning using noisy and uncertain real-world data. 
\cite{poon2006sound} introduce statistical sampling to improve the efficiency of search for satisfiability testing.
\cite{richardson2006markov,singla2012markov,poon2009unsupervised,raedt2008probabilistic} introduce Markov logic networks, and
form the joint distribution of a probabilistic graphical model by weighting the formulas in a first-order logic. 

Our approach shares with Markov logic networks the philosophy of combining logical 
tools with probabilistic modeling. Markov logic networks utilize a general first-order logic to help infer relationships among objects, but they do not 
explicitly address the planning problem, and we note the formulation and solution of planning problems 
in first-order logic is often inefficient.
Our approach instead exploits the highly structured planning domain 
by integrating a widely used logical plan validator within the probabilistic generative model.


\section{Algorithm}\label{sec:algorithm}

\subsection{Generative Model}
\begin{figure}
\begin{center}
\includegraphics[scale=0.8]{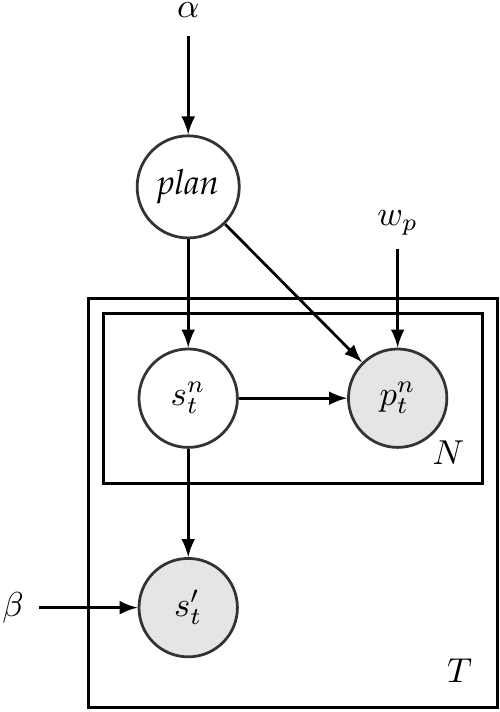}
\caption{The generative model\label{fig:graphicalModel}}
\end{center}
\end{figure}

In this section, we explain the generative model of human team planning dialog (Figure~\ref{fig:graphicalModel})
that we use to infer the team's final plan.
We start with a $plan$ latent variable that 
must be inferred by observing utterances in the planning session. 
The model generates each utterance in the conversation by sampling a subset of the predicates in $plan$ and computing the relative ordering in which they appear in the utterance.
This mapping from the absolute ordering in $plan$ to the relative ordering of predicates in an utterance is described in more detail below. 
Since the conversation is short, and the noise level is high, our model does not distinguish utterances based on the order of which they appear in the conversation.

We describe the generative model step by step:

\begin{enumerate}
    \item \textbf{Variable $plan$:} The $plan$ variable in Figure~\ref{fig:graphicalModel} 
    is defined as an ordered tuple of sets of grounded predicates and 
    represents
the final plan agreed upon by the team. 
The prior distribution over the $plan$ variable is given by:
\begin{equation}
\label{eq:plandistr}
 p(plan) 
 \propto \begin{cases} e^{\alpha} & \text{if plan is valid}\\
     1 & \text{if plan is invalid.}  
    \end{cases}
\end{equation}

 where $\alpha$ is a positive number. This models our assumption that the final plan is more likely,
 but not necessarily required, to be a valid plan. 
 We describe in the next section how we evaluate the validity of a plan. 

Each set of predicates in $plan$ is assigned a consecutive 
absolute plan step index $s$, starting at $s=1$  working from left to right in the ordered tuple.
For example, given $plan=(\{A_1, A_2\},\{A_3\}, \{A_4,A_5,A_6\})$, 
where each $A_i$ is a predicate, $A_1$ and $A_2$ occur in parallel at plan step $s=1$ and $A_6$ occurs at plan step $s=3$.

\item \textbf{Variable $\boldsymbol s_t^n$:} Each of $n$ predicates
in an utterance $t$ is assigned a step index $s_t^n$.
$s_t^n$ is given by the \emph{absolute} plan step $s$ where the corresponding predicate $p_t^n$ (explained later), appears in $plan$.

$s_t^n$ is sampled as follows.  
For each utterance, $n$ predicates are sampled from $plan$. 
For example, consider $n=2$ where first sampled predicate appears in the second set of $plan$ and the second sampled predicate appears in the fourth set of $plan$. Then $s_t^1=2$ and $s_t^2=4$.
The probability of a set being sampled is proportional to the number of predicates in the set.
For example, given $plan = (\{A_1, A_2\},\{A_3\}, \{A_4,A_5,A_6\})$,
the probability of selecting the first set ($\{A_1, A_2\}$)
is $\frac{2}{6}$. This models that people are more likely to discuss plan steps that include
many predicates. Formally,

\begin{align}
\label{eq:stndistr}
p(s_t^n = i | plan)
= \frac{\text{\# predicates in set $i$ in plan}}{\text{total \# predicates in plan}}.
\end{align}

\item \textbf{Variable $\boldsymbol s'_t$:} The variable $s'_t$ is an array of size $n$ that specifies, for each utterance $t$, the \emph{relative} ordering of predicates as they appear in $plan$. 

$s'_t$ is generated from $s_t$ as follows:
\begin{equation}
 \label{eq:sprimedistr}
p(s'_t|s_t) 
\propto
\begin{cases} e^{\beta} & \text{ if }  s'_t = f(s_t)  \\ 
1 & \text{ if } s'_t \neq f(s_t).
\end{cases}
\end{equation}

where $\beta>0$. The function $f$ is a deterministic mapping from the absolute ordering $s_t$ to the relative ordering $s'_t$.
$f$ takes as input a vector of absolute plan step indices and produces
a vector of consecutive indeces. For example, $f$ maps
$s_t=(2,4)$ to $s'_t=(1,2)$, and $s_t=(5,7,2)$ to $s'_t=(2,3,1)$.
 
This variable models the way predicates and their orders appear in human conversation; people
do not often refer to the absolute ordering, and frequently use relative terms such as ``before'' and ``after'' to describe 
partial sequences of the full plan. People also make mistakes or otherwise incorrectly specify an ordering, hence our model allows 
for inconsistent relative orderings with nonzero probability.

\item \textbf{Variable $\boldsymbol p_t^n$:} The variable $p_t^n$ represents
the $n$th predicate that appears in the $t$th utterance. This is sampled given $s_t^n$, the $plan$ variable and a noise parameter $w_p$.

With probability $w_p$, we sample the predicate $p_t^n$ uniformly from the
``correct'' set $s_t^n$ in $plan$ as follows:
\begin{align*}
p(p_t^n = i &| plan, s_t^n=j) 
    = \begin{cases}
        \frac{1}{\text{\# pred. in set } j} & \text{if $i$ is in set $j$}\\
        0 &\text{o.w.}
    \end{cases}
\end{align*}

With probability $1-w_p$, we sample the predicate $p_t^n$ uniformly from 
``any'' set in $plan$ (i.e.~from all predicates mentioned in the dialog), therefore 
\begin{align}
\label{eq:ptndistr}
p(&p_t^n = i | plan, s_t^n=j) 
= \frac{1}{\text{total \# predicates}}.
\end{align}
In other words, with higher probability $w_p$,
we sample a value for $p_t^n$ that is consistent with $s_t^n$, but allow nonzero probability that $p_t^n$ is sampled
from a random plan. This allows the model to incorporate the noise in the planning conversation, including mistakes or plans that are later revised.

\end{enumerate}

\subsection{PDDL Validator}
We use the Planning Domain Description Language (PDDL) 2.1 plan validator tool~\cite{howey2004val} to evaluate the prior distribution over possible plans. 
The plan validator is a standard tool that takes as input a planning problem described in PDDL and a proposed solution plan, and checks if the plan is valid. 
Some validators can also compute a value for plan quality, if the problem specification includes a plan metric. 
Validation is often much cheaper than generating a valid plan, and gives us a way to compute $p(plan)$ up to proportionality in a computationally efficient manner. 
Leveraging this efficiency, we use Metropolis-Hastings sampling to sample the plan without calculating the partition function, as described in the next section.

Many elements of the PDDL domain and problem specification are defined by the capabilities and resources at a particular response team's disposal. The PDDL specification may largely be reused from
mission to mission, but some elements will change or may be specified incorrectly. For example the plan specification may misrepresent the number of responders available or may leave out implicit constraints that people take for granted. 
In the Experimental Evaluation section we demonstrate the robustness of our approach using both complete and degraded PDDL plan specifications.

\subsection{Gibbs Sampling~\label{sec:gibbs}}
We use Gibbs sampling to perform inference on the generative model. 
There are two latent variables to sample: $plan$ and the collection of variables $s_t^n$.
We iterate between sampling $plan$ given all other variables and 
sampling the $s_t^n$ variables given all other variables. The PDDL
validator is used when the $plan$ variable is sampled. 

Unlike $s_t^n$, where we can write down an analytic form to sample
from the posterior, it is intractable to directly resample the $plan$ variable, as
it requires calculating the number of all possible valid and invalid plans. 
Therefore we use a Metropolis-Hasting (MH) algorithm to sample from the
$plan$ posterior distribution within the Gibbs sampling steps.
In this section, we explain how $plan$ and the $s_t^n$ variables are sampled.

\subsubsection{Sampling plan using Metropolis-Hastings\label{sec:samplePlan}}
The posterior of $plan$ can be represented as the product of the prior 
and likelihood as follows:
\begin{align}
\label{eq:samplePlan}
 p(plan|s, p) 
 &\propto  p(plan) p(s,p|plan) \nonumber \\
 &= p(plan)\prod_{t=1}^T \prod_{n=1}^N  p(s_t^n, p_t^n | plan) \nonumber \\
 &= p(plan)\prod_{t=1}^T \prod_{n=1}^N  p(s_t^n|plan) p(p_t^n| plan, s_t^n) 
\end{align}


The MH sampling algorithm is widely used to sample from a distribution when direct sampling
is difficult. The typical MH algorithm defines a proposal distribution, $Q(x'|x_t)$ 
which samples a new point (i.e.~$x'$: a value of the $plan$ variable in our case) given the
current point $x_t$. The new point
can be achieved by randomly selecting one of the possible moves, as defined below.
The proposed point is
accepted or rejected with probability of $\min (1,\text{acceptance ratio})$. 

Unlike simple cases where a Gaussian distribution can be used as a proposal distribution, 
our proposal distribution needs to be defined over the plan space. 
Recall that $plan$ is represented as an ordered tuple
of sets of predicates, and the new point is suggested by performing one of the following moves:
\begin{itemize}
    \item Select a predicate. If it is in the current $plan$, move it  
        to either: 1) the next set of predicates, 2) the previous set or 3) remove
        it from the current $plan$. If it is not in the current $plan$, move it to one of the existing set. 
\item Select two sets in $plan$, and switch their orders. 
\end{itemize}
These moves are sufficient to move from an arbitrary plan to another arbitrary plan
using the set of moves defined above.
The probability of each of these moves described above is chosen such that
the  the proposal distribution is symmetric: $Q(x'|x_t) = Q(x_t|x')$.
This symmetry simplifies the calculation of the acceptance ratio.

Second, the ratios of the proposal distribution at the current and proposed points are calculated. 
When $plan$ is valid and invalid respectively, $p(plan)$ is proportional to $e^{\alpha}$
and 1 respectively, as described in Equation~\ref{eq:plandistr}. Plan validity is evaluated using the PDDL validation tool.
The remaining term, $p(s_t^n|plan) p(p_t^n| plan, s_t^n)$,
is calculated using Equations~\ref{eq:stndistr} and~\ref{eq:ptndistr}.

Third, the proposed $plan$ is accepted with the following probability:
$\min \left(1, \frac{ p^*(plan=x'|s,p) }{ p^*(plan=x_t|s,p) }\right)$, where $p^*$ is a function that is proportional 
to the posterior distribution.

\subsubsection{Sampling $s_t^n$}
Fortunately, there exists an analytic expression for the posterior of $s_t^n$:
\begin{align*}
    p(s_t | plan, p_t, s'_t)
                             & \propto p(s_t|plan) p(p_t, s'_t|plan,  s_t)\\
                             & = p(s_t|plan) p(p_t|plan, s_t ) p(s'_t|s_t)\\
                             & = p(s'_t|s_t) \prod_{n=1}^N p(s_t^n|plan) p(p_t^n|plan, s_t^n)
\end{align*}
 
Note this analytic expression can be expensive to evaluate if the number of possible values of $s_t^n$ is large. In that case
one can marginalize out $s_t^n$, as the one variable we truly care about is the $plan$ variable.


\section{Experimental Evaluation}

In this section we evaluate the performance of our plan inference algorithm 
through initial proof-of-concept human subject experimentation and show we are able to infer a human team's final plan with 83\% accuracy on average. We also describe a 
robot demonstration in which two people plan and execute a first-response collaborative task with a PR2 robot.

\noindent\textbf{Human Team Planning Data} 
We designed a web-based collaboration tool that is modeled after the NICS system~\cite{LincolnNICS} used by first response teams, but with a modification that requires the team to communicate soley via text chat. 
Twenty-three teams of two (total of 46 participants) were recruited through Amazon Mechanical Turk and the greater Boston area. Recruitment was restricted to participants located in US 
to increase the probability that participants were fluent in English. Each team was provided the fictional 
rescue scenario described in this paper, and was asked to collaboratively plan a rescue mission.
At the completion of the planning session, each participant was asked to summarize the final agreed upon plan in the structured form described previously. 
An independent analyst reviewed the planning sessions to resolve discrepancies between the two member's final plan descriptions, when necessary. 
Utterance tagging was performed by the three analysts: the first and second authors, and an independent analyst. Two of the three analysts tagged and reviewed
each team planning session. On average, 19\% of predicates mentioned per data set did not end up in the final plan. 
 
\noindent\textbf{Algorithm Implementation} The algorithm is implemented in Python, and the VAL PDDL 2.1 plan validator~\cite{howey2004val} is used.  
We perform 3000 Gibbs sampling steps on the data from each planning session. Within
one Gibbs sampling step, we perform 400 steps of the Metropolis-Hastings (MH) algorithm 
to sample the plan. Every 20 samples are selected to measure the accuracy.
$w_p$ is set to $0.8$, $\alpha$ is 10, $\beta$ is 5.
 
\noindent\textbf{Results} We evaluate the quality of the final plan produced by our algorithm in terms of (1) accuracy of task allocation among agents (e.g. which medic travels to which room), and
(2) accuracy of plan sequence. 

Two metrics for task allocation accuracy are evaluated: 1) [\% Inferred] the percent of inferred plan predicates that appear in the team's final plan, 
and 2) [\% Noise Rej] the percent noise rejection of extraneous predicates that are discussed but do not appear in the team's final plan.   

We evaluate the accuracy of the plan sequence as follows.  
We say a pair of 
predicates is \emph{correctly ordered} if it is consistent with the relative ordering in the true final plan.
We measure the percent accuracy of sequencing [\% Seq] by
$\frac{\text{\# correctly ordered pairs of correct predicates}}
{\text{total \# of pairs of correct predicates}}$.
Only correctly estimated
predicates are compared, as there is no ground truth relation for predicates that are not in the true final 
plan. 
We use this relative sequencing measure because it does not compound the sequence errors, as an absolute difference measure would  
(e.g. where the error in the ordering of one predicate early in the plan shifts the position of all subsequent predicates).
 
Overall plan accuracy is computed as the arithmetic mean of the two task allocation and one plan sequence accuracy measures. 
Our algorithm is evaluated under three conditions: 1) [PDDL] perfect PDDL files, 2) [PDDL-1] PDDL problem file with missing
goals/constants (delete one patient and one robot), 3) [PDDL-2] PDDL domain file missing a constraint (delete precondition that a
robot needs to inspect the room before a human crew can enter), and 4) [no PDDL] using an uninformative prior over possible plans.

Results, shown in Table~\ref{tab:result}, show that our algorithm infers final plans with greater than 80\% on average, and achieves 
good noise rejection of extraneous predicates discussed in conversation. We also show our approach is relatively robust to degraded PDDL
specifications.

\begin{table}
\begin{center}
\small
 \begin{tabular}{|p{1.4cm}|| c|c||c|| c|}
  \hline 
& \multicolumn{2}{c||} {Task Allocation }& \multirow{2}{*}{\% Seq} & \multirow{2}{*}{Avg.}\\ \cline{1-3}
  \parbox{2cm}{ }& \% Inferred & \% Noise Rej &  &\\
\hline
 PDDL			 & 92.2 & 71 & 85.8 &  83.0\\ \hline
 \parbox{2cm}{PDDL-1}    & 91.4 & 67.9 & 84.1 & 81.1\\\hline
 \parbox{2cm} {PDDL-2}   & 90.8 & 68.5  & 84.2 & 81.1 \\ \hline
 No PDDL 		 & 88.8 & 68.2 & 73.4 & 76.8\\ \hline
 \end{tabular}
\end{center}
\caption{Plan Accuracy Results\label{tab:result}}
\end{table}

\noindent\textbf{Concept-of-Operations Robot Demonstration}  
We illustrate the use of our plan inference algorithm with a 
robot demonstration in which two people plan and execute a first-response collaborative task with a PR2 robot.
Two participants plan an impending deployment using the web-based collaborative tool that we developed. Once the planning session is complete,
the dialog is manually tagged. 
The plan inferred from this data is confirmed with the human planners, and is provided to the 
robot to execute. The registration of predicates to robot actions, and room names to map locations, is performed offline in advance. 
While the first responders are on their way to the accident scene, the PR2
autonomously navigates to each room performing online localization, path planning, and obstacle avoidance. The 
robot informs the rest of the team as it inspects each room and confirms it is safe for the human team members to enter. 
Video of this demo is here: \texttt{http://tiny.cc/uxhcrw}.

\section{Future Work}

Our aim is to reduce the burden
of programming a robot to work in concert with a human team. We present an algorithm that combines a probabilistic approach with logical plan validation
to infer a plan from human team conversation. We empirically demonstrate that this hybrid approach
enables us to infer 
the team's final plan with about 83\% accuracy on average.

Next, we plan to investigate the speed and ease with which a human planner can ``fix'' the inferred plan to achieve near 100\% plan accuracy. 
We are also investigating automatic mechanisms for translating raw natural dialog into the structured form our algorithm takes as input. 
 
\section{Acknowledgement}

The authors thank Matt Johnson and James Saunderson for helpful discussions and feedback.

This work is sponsored by ASD (R\&E) under Air Force Contract
FA8721-05-C-0002. Opinions, interpretations, conclusions and recommendations
are those of the authors and are not necessarily endorsed by the United
States Government. 

 \bibliographystyle{aaai}

\bibliography{mybib}

\end{document}